# Time manipulation technique for speeding up reinforcement learning in simulations

Petar Kormushev[1], Kohei Nomoto[2], Fangyan Dong[3], Kaoru Hirota[4]

[1, 2, 3, 4] Department of Computational Intelligence and Systems Science
Tokyo Institute of Technology, Japan
E-mails: {petar, nomoto, tou, hirota}@hrt.dis.titech.ac.jp

[2] Industrial Design Center, Mitsubishi Electric Corporation, Japan
E-mail: Nomoto.Kohei@dw.MitsubishiElectric.co.jp

*Abstract:* A technique for speeding up reinforcement learning algorithms by using time manipulation is proposed. It is applicable to failure-avoidance control problems running in a computer simulation. Turning the time of the simulation backwards on failure events is shown to speed up the learning by 260% and improve the state space exploration by 12% on the cart-pole balancing task, compared to the conventional Q-learning and Actor-Critic algorithms.

1. Introduction

Reinforcement Learning (RL) algorithms have been applied successfully for many years [4]. One of their main virtues is that they don't require a model of the device they are supposed to control [7]. Also, general RL algorithms like Q-learning [10] and TD($\lambda$) [8] provably converge to the globally optimal solution (under some assumptions) [1, 10].

This convenience, however, comes at a certain cost. The price for this flexibility of RL algorithms is that they require long training [9]. Even for a relatively simple control task (e.g. the inverted pendulum balancing problem, also known as the cart-pole balancing problem [2]) a general RL algorithm requires many trials (hundreds or even thousands of trials) to be able to learn the task.

What is the reason for this slow convergence? Trying to answer this question, let us focus on a subset of the RL problems: learning a control policy to avoid failure. The inverted pendulum balancing task is a good representative example for such a problem. In this case, failure is defined as falling of the pole beyond a certain angle from the upright position or hitting the edges of the cart track. The aim of the RL algorithm is to find a control policy which can prevent the pendulum from falling by moving the cart, without hitting the edges of the cart track. A good name for such problems is "failure-avoidance problems". They follow the general "trial-and-error" paradigm of unsupervised learning [11].

The learning process is organized in separate trials (or episodes), each starting from the same initial position at the center of the cart track and finishing in a failure state (either when the pendulum has tilted too much, or when the cart has hit an edge). After every failure, the state of the pendulum is reset back to the initial position and the next trial begins.

Usually, the reward function in such problems is defined as -1 in case of a failure and 0 in all other cases. Thus, the RL agent is trying to maximize the cumulative reward effectively avoiding failure states. The failure states include tilting the pendulum more than a certain angle, as well as hitting the left or right edges of the cart track. The target is, for example, to keep the pendulum balanced for at least 100 000 steps in a single trial.

For this particular problem, a general RL algorithm like Q-learning will need more than 1000 trials and more than 200 000 total steps to reach the target. Why does it take so long?

One main reason for this is the poor state space exploration. Practically all RL algorithms follow the same scheme of doing trials. Each time a failure occurs, a new trial begins from the initial state. As a consequence, the state space close to the initial state is very well explored, but the state space further from the initial state is not. In this particular example, the RL algorithm learns to balance the pendulum around the initial position very quickly, but as the pendulum goes further and reaches the ends of the track, it fails immediately. And this is completely understandable, since the RL agent doesn't have enough "experience" in that part of the state space.

The main problem under investigation here is: how to improve the state space exploration, providing enough "experience" for the RL agent in a broader part of the state space. For example, instead of exploring the same state space near the initial state over and over again, it is desirable for the RL algorithm to focus on the states which lead to failure and try harder to avoid them. This would improve dramatically the state space exploration and speed up the learning process. The present paper proposes a time manipulation technique to achieve this goal.

The key idea is that it is possible to manipulate the time of the simulation in such a way which forces the RL algorithm to explore better the state space in proximity to failures. At the same time, it is possible to avoid re-visiting already well-explored parts of the state space. The time manipulation, which is proposed, consists of turning the time of the simulation backwards when failure events occur, while at the same time preserving the learned policy as it was at the time of the failure. This technique is shown to improve substantially the learning speed and the state space exploration of Q-learning and Actor-Critic algorithms, at the expense of using additional memory. Also, it has the advantage of being completely transparent for the RL algorithm.

Section 2 describes the general RL algorithm for solving failure-avoidance problems. Section 3 is devoted to the proposed time manipulation technique for improving the previously mentioned algorithm. Section 4 describes the experimental evaluation of the proposed technique on a classical benchmark RL problem.

## 2. The general RL algorithm for solving failure-avoidance problems

For the purpose of explaining the proposed time manipulation technique any general RL algorithm for solving failure-avoidance problems is usable. Probably the most widely known and used such algorithm is Q-learning [10]. It is a classical form of RL algorithm that does not need a model of its environment and can be used for solving failure-avoidance problems. Because of its popularity, Q-learning was selected as a very good candidate for describing and testing the proposed time manipulation technique. This section describes the standard Q-learning algorithm and the next section explains how to modify it using the proposed time manipulation technique.

The Q-learning algorithm works by estimating the values of state-action pairs. A lookup table is used to store the Q-values. The value Q($s$, $a$) is defined to be the expected discounted sum of future rewards obtained by taking action $a$ from state $s$ and following an optimal policy thereafter. Once these values have been learned, the optimal action from any state is the one with the highest Q-value. The algorithm to estimate the Q-values based on experience is as follows:

| Standard Q-learning algorithm for RL |
| --- |
| 1) Initialization<br>    Set all Q-values to zero.<br>    Set the current state $s_t$ to be the initial state $s_0$.<br>2) From the current state $s_t$, select an action $a_t$. This will cause a receipt of an immediate reward $r_t$, and arrival at a next state $s_{t+1}$.<br>3) Update Q($s_t$, $a_t$) based on the acquired experience as follows:<br>    Q($s_t$, $a_t$) ← Q($s_t$, $a_t$) + $\alpha_t(s_t, a_t)$[$r_t$ + $\gamma$ max$_a$Q($s_{t+1}$, $a$) − Q($s_t$, $a_t$)]<br>    where $0 \leq \alpha_t(s_t, a_t) \leq 1$ is the learning rate and $0 \leq \gamma < 1$ is the discount factor<br>4) Go to step 2. |

This algorithm is guaranteed to converge to the correct Q-values with probability 1 if the following conditions are satisfied: the environment is stationary and Markovian (depends on the current state and the action taken in it only), every state-action pair continues to be visited, and the learning rate is decreased appropriately over time [10]. This exploration strategy does not specify which action to select at each step. Usually, different methods for selecting actions are used (such as $\varepsilon$-greedy or Boltzmann distribution strategy [3]), which ensure that all actions have chance of being selected while still favoring actions with higher value estimates.

The standard Q-learning algorithm, however, cannot be used directly in practice, because most of the problems require keeping the state within some set of allowed or desired states. Therefore, the learning process is organized in trials and every time when the state changes to an undesirable state (failure state), a new trial starts from the initial state $s_0$. This is the resulting algorithm:

| Standard Q-Learning algorithm with trial resetting on failure |
|---|
| 1) Initialization<br>    Set all Q-values to zero.<br>    Set the current state $s_t$ to be the initial state $s_0$.<br>2) From the current state $s_t$, select an action $a_t$. This will cause a receipt of an immediate reward $r_t$, and arrival at a next state $s_{t+1}$. The reward $r_t$ is defined to be -1 if $s_{t+1}$ is a failure state.<br>3) Update $Q(s_t, a_t)$ based on the acquired experience as follows:<br>    $Q(s_t, a_t) \leftarrow Q(s_t, a_t) + \alpha_t(s_t, a_t)[r_t + \gamma \max_a Q(s_{t+1}, a) - Q(s_t, a_t)]$<br>    where $0 \leq \alpha_t(s_t, a_t) \leq 1$ is the learning rate and $0 \leq \gamma < 1$ is the discount factor<br>4) If the new state $s_{t+1}$ is a failure state, mark the current trial as a failure and reset the current state $s_t$ to the initial state $s_0$.<br>5) Go to step 2. |

Due to the trial resetting on failure events, this algorithm has some major disadvantages:
- The state space around the initial state becomes very well explored (because every trial starts from it), but not any further states, especially states close to failure states.
- A lot of computational power is used just to repeat already explored states and state transitions, which makes the Q-values change slowly over time, thus slowing down the learning progress. The longer the algorithm is running, the slower it learns new experience.
- Hundreds and even thousands of trials are required to be able to learn a longer sequence of state transitions, in order to avoid failure states for longer time.

Trying to overcome the mentioned disadvantages, a novel time manipulation technique to improve the standard RL algorithm with trial resetting on failure is proposed, as described in section 3.

## 3. Time manipulation technique

The proposed time manipulation technique is in essence a modification of the standard RL algorithm presented in section 2. Before giving the implementation details, let us start with an intuitive description of the basic idea.

RL algorithms are very useful when trying to teach a robot to perform some motor control task. In most cases however, it is not practical to run the RL algorithms directly on the physical robot, because it may break during the trials, or because it takes too much time and resources. This is especially true for failure-avoidance problems, since failure states are often associated with falling or breaking of the robot, which is not desirable. In such cases, computer simulations are used first, and later the learned policy is fine-tuned on the physical robot.

Virtual simulations have many advantages over real-world experiments. One such prominent advantage is the control over time. For example, you can speed up or slow down the time inside a simulation. You can even make the time go backwards and go back to events that already happened. Also, you can repeat exactly the same experiment twice using exactly the same environment. Such time manipulations are impossible in the real-world, but this paper is going to show that they can be extremely beneficial for a RL algorithm running inside a simulation. The proposed time manipulation technique helps to speed up learning and explore the state space better.

The main focus of the proposed idea is the failure event which occurs during the RL simulation. As mentioned in section 2, when a failure state is incurred, the standard RL algorithm immediately stops the current trial and resets the state to the initial one. However, it is exactly the state space in proximity to the failure states which is "interesting" for the RL algorithm and it should be better explored instead of abandoned so quickly.

Of course, normally, after failure there is no action that can be done to continue the trial. This is where the power of virtual simulations comes into play. It is actually possible to manipulate the time of the simulation and turn it back to a previous point in time, shortly before the failure occurred. Only this is not enough, however, because the RL algorithm is going to choose the same actions and the simulation will repeat itself. To avoid this, while turning back the time, it is necessary to retain the newly learned experience of the failure by preserving the RL policy (the Q-values in the case of Q-learning). This time, the RL algorithm will try to avoid the failure state by choosing other actions shortly before it. The modified Q-learning algorithm looks like this:

| Time Manipulated Q-learning algorithm |
|---|
| 1) Initialization<br>      Set all Q-values to zero.<br>      Set the current state $s_t$ to be the initial state $s_0$.<br>      Set the sequence of saved states to be $<s_0>$ |
| 2) From the current state $s_t$, select an action $a_t$. This will cause a receipt of an immediate reward $r_t$, and arrival at a next state $s_{t+1}$. The reward $r_t$ is defined to be -1 if $s_{t+1}$ is a failure state. |
| 3) Add the new state $s_{t+1}$ to the sequence of saved states $<s_0, s_1, s_2, \ldots>$ |
| 4) Update $Q(s_t, a_t)$ based on the acquired experience as follows:<br>      $Q(s_t, a_t) \leftarrow Q(s_t, a_t) + \alpha_t(s_t, a_t)[r_t + \gamma \max_a Q(s_{t+1}, a) - Q(s_t, a_t)]$<br>      where $0 \leq \alpha_t(s_t, a_t) \leq 1$ is the learning rate and $0 \leq \gamma < 1$ is the discount factor |
| 5) If the new state $s_{t+1}$ is a failure state, turn back the time of the simulation to the previous saved state $s_t$ in the time sequence, preserving the current Q-values. |
| 6) Go to step 2. |

What if choosing other actions does not help and failure occurs again? In this case, the time has to be turned even further back in the simulation, earlier enough so that the RL agent can perform actions to avoid the failure. The absolute extreme is going back to the initial state of the trial, which is identical to trial resetting in the standard RL algorithm. The minimum is going back just 1 step before the failure occurred. For practical purposes, going back to the middle of the failed trial seems to provide a good balance between these two extremes. This strategy was employed during the experiments described in section 4 and it showed satisfactory results. Another approach is to select probabilistically a previous state by using, for example, some distribution strategy.

It is interesting to mention that, actually, using this technique it might be possible to finish learning in only one trial. If the RL agent is doing only "small" time manipulations to recover from failures, there is no need to start a new trial at all. This means that the RL agent could learn in only one single trial (ignoring the small time manipulations).

Another important issue to deal with are the so called "eligibility traces" [6, 7]. Eligibility traces are one of the basic mechanisms for temporal credit assignment in reinforcement learning. An eligibility trace is a temporary record of the occurrence of an event, such as the visiting of a state or the taking of an action. For example, in the popular temporal-difference algorithm TD($\lambda$), the $\lambda$ refers to the use of an eligibility trace. The trace marks the memory parameters associated with the event as eligible for undergoing learning changes. When a policy update occurs, only the eligible states or actions are assigned credit or blame for the received reward. Almost any temporal-difference method, e.g. Q-learning or Sarsa [5], can be combined with eligibility traces to obtain a more general method that may learn more efficiently. [7]

The question is, if eligibility traces are used is it still possible to apply the proposed time manipulation technique? And the answer is affirmative, providing that care is taken to adjust the values of the eligibility traces correctly when turning back the time of the simulation. In the case when the eligibility traces $e_t(s)$ are decayed by a constant rate $\lambda$ at each time step, the forward update looks like this:

$$e_t(s) = \begin{cases} \lambda \gamma\, e_{t-1}(s) & \text{if } s \neq s_t; \\ \lambda \gamma\, e_{t-1}(s) + 1 & \text{if } s = s_t, \end{cases} \quad (1)$$

where $0 \leq \lambda < 1$ is the decay rate for the eligibility traces and $0 \leq \gamma < 1$ is the discount factor. In order to reverse the eligibility traces one step back in time, the following backward update is required:

$$e_{t-1}(s) = \begin{cases} e_t(s) / (\lambda\gamma) & \text{if } s \neq s_t; \\ (e_t(s) - 1) / (\lambda\gamma) & \text{if } s = s_t. \end{cases} \quad (2)$$

To conclude the description of the proposed time manipulation technique, let us analyze it from computational complexity viewpoint. Obviously, the technique is trying to improve the speed of learning at the expense of using more memory. The algorithm needs enough memory to store the complete state of the simulation for every state for which it is required to be able to go back to. In this sense, it is possible to implement efficiently the proposed technique in such a way that it uses only a limited amount of memory. This can be done by deleting some of the old saved states, e.g. every alternating state, in order to reduce the memory consumption while still keeping the ability to turn back in time. The need to do this will occur only in case of very long trials and when the state variables are extremely numerous. In such case the used memory can be adapted to the available resources dynamically by limiting the number of saved states during execution.

An important advantage of the proposed time manipulation technique is its complete transparency for the RL algorithm, in the sense that the RL algorithm does not even realize that the time of the simulation is being manipulated externally. From the viewpoint of the RL algorithm, the time is flowing only forward and all the trials are just an ordered sequence of states in a linearly flowing time (this is, of course, providing that the time manipulation takes care to save and restore the appropriate eligibility traces along with the rest of the simulation parameters), whereas, in fact, the sequence is a tree of time interconnected state sequences. Being transparent for the RL algorithm means that the proposed technique can easily be combined with arbitrary RL algorithm, improving its performance without changing the algorithm's logic at all.

# 4. Evaluation of the learning speed and the state space exploration

The proposed time manipulation technique was extensively tested using different conventional RL algorithms, various parameters for the time manipulation technique and variations of the problem parameters. This section presents the experimental environment, the conducted experiments and the obtained results.

## 4.1. The TiME experimental environment

A dedicated experimental system called TiME (**Ti**me **M**anipulation **E**nvironment) was developed especially for the purpose of this evaluation. A general view of the environment is shown on Figure 1. TiME has a built-in simulation engine, implementation of two conventional RL algorithms (Q-learning and Actor-critic algorithm), useful visualization modules (for the simulation, the learning data and the state transitions graph) and most importantly – a prototype implementation of the time manipulation technique. To facilitate the analysis of the algorithm behavior, TiME displays detailed information about the current state, the previous state transitions, a visual view of the simulation, and allows runtime modification of all important parameters of the algorithms and the simulation. There is a manual and automatic control of the time manipulation technique, as well as visualization of the accumulated data in the form of charts.

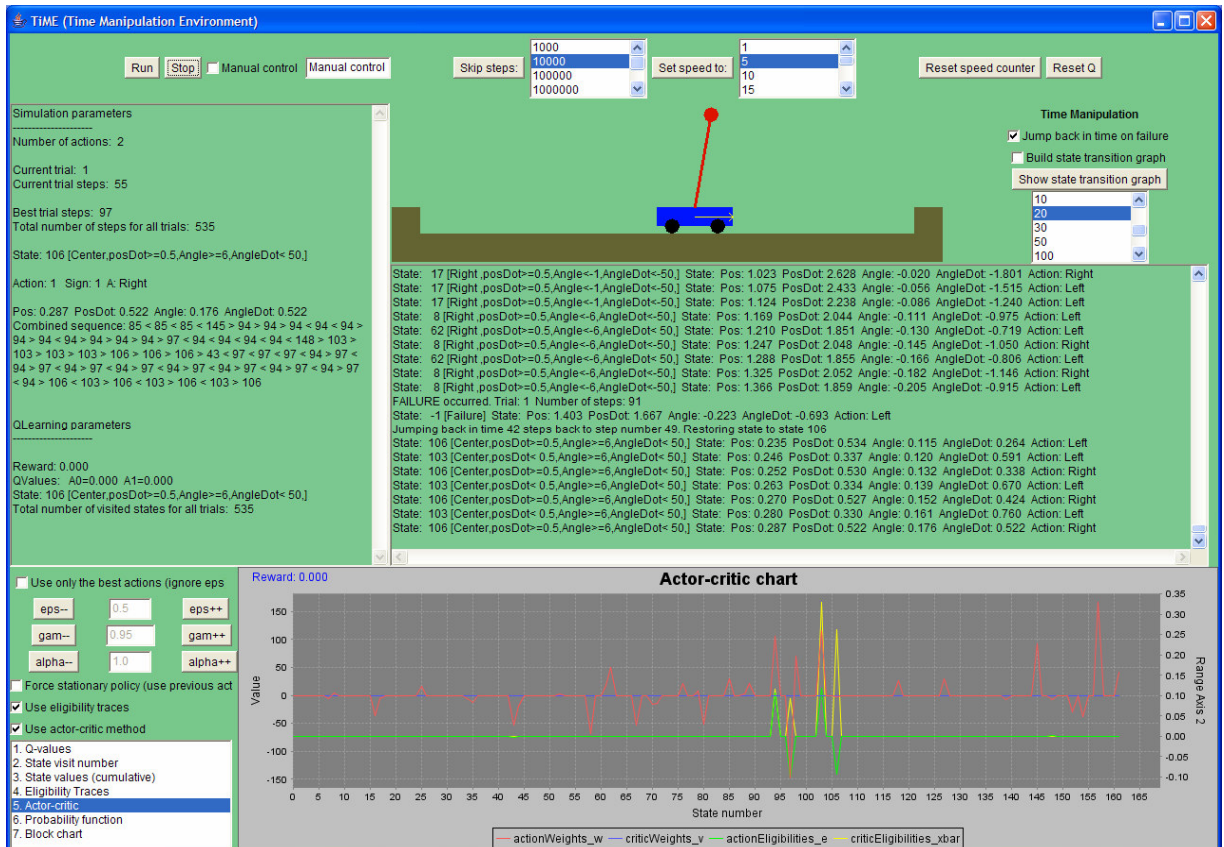

Figure 1. General view of TiME (**Ti**me **M**anipulation **E**nvironment).

The experiments were conducted on a classical failure-avoidance RL problem: the cart-pole balancing task. This problem was chosen because it is very well studied and is commonly used as a benchmark for new RL algorithms. During the experiments, failure was defined as falling of the pole beyond a 12-degree angle from the upright position or hitting the edges of the cart track. The reward function was defined as -1 in case of a failure and 0 in all other cases. The values of the physical parameters used in the simulation are given in Table 1. A fixed force magnitude was used for both left and right push of the cart.

| Physical parameter of the simulation | Value [units] |
|---|---|
| Length of the cart track | 4.4 [m] |
| Length of the pole | 1.0 [m] |
| Mass of the pole | 0.1 [kg] |
| Mass of the cart | 1.0 [kg] |
| Simulation time slice duration | 0.02 [sec] |
| Force magnitude for pushing the cart | 10.0 [N] |

Table 1. The values of the physical parameters used during the experiments.

Although the proposed time manipulation technique is not a stand-alone RL algorithm, it can easily be used to modify any general RL algorithm to produce a time manipulated version of it. For the experiments, two conventional RL algorithms were used: Q-learning and Actor-Critic learning. During the preliminary tests with the unmodified versions of the conventional algorithms, it was found that the Actor-Critic method is performing much better on this task than the Q-learning. This is probably due to the fact that the Actor-Critic method maintains separate representations for the action selection part and the state evaluation part of the algorithm. The results presented in this section compare the time manipulation technique with the better one of the two conventional RL algorithms: the Actor-Critic method. Otherwise, the implementation of the time manipulation technique is the same for both Q-learning and Actor-Critic method.

The conducted experiments can be classified in two groups: experiments to evaluate the technique's effect on the speed of learning and on the exploration of the state space. The experiments from both groups were conducted in a similar way. First, the unmodified RL algorithm was used with a set of different algorithm parameters and problem settings. After that, the time manipulation technique was added to the algorithm, and the same set of parameters and settings was used in exactly the same simulation environment and initial state. This schema was repeated 10 times and the results were averaged. The unmodified RL algorithm and the time manipulated version of it were compared based on the two sets of results.

4.2. Evaluation of the technique's effect on the learning speed

The two versions (with and without the time manipulation) were trained for the same number of steps. During the training the best (longest) trial steps were recorded, as well as the number of unique states visited. The training was stopped (the running trial was interrupted if necessary) and a benchmark trial was executed. This schema was repeated 12 times while increasing the total number of training steps. Table 2 shows the experimental results.

| Total training steps | Best trial steps | | Benchmark trial steps | | Unique states visited | |
|---|---|---|---|---|---|---|
| | Without time manipulation | With time manipulation | Without time manipulation | With time manipulation | Without time manipulation | With time manipulation |
| 100 | 25 | 66 | 32 | 48 | 18 | 20 |
| 200 | 57 | 75 | 64 | 71 | 26 | 26 |
| 500 | 92 | 147 | 111 | 113 | 31 | 32 |
| 1000 | 160 | 165 | 114 | 162 | 43 | 51 |
| 2000 | 187 | 350 | 103 | 311 | 54 | 63 |
| 5000 | 435 | 732 | 255 | 319 | 76 | 87 |
| 10000 | 1053 | 3608 | 867 | 4280 | 91 | 96 |
| 20000 | 3245 | 6192 | 2476 | 19855 | 99 | 113 |
| 30000 | 4707 | 6412 | 7534 | 69345 | 104 | 116 |
| 40000 | 5381 | 12561 | 25854 | 134032 | 101 | 114 |
| 50000 | 12341 | 21078 | 89722 | 267621 | 104 | 121 |
| 100000 | 28163 | 68523 | over 500000 | over 500000 | 108 | 125 |

Table 2. The averaged experimental results for 12 groups of experiments.

The difference in the learning speed of the two versions can be seen by comparing the duration of the best trials and the benchmark trials after each training set. Figure 2 shows the comparison results. During all the tests, the time manipulation technique achieved better results: longer best trials, as well as longer benchmark trials for the same amount of training steps. The reason for longer best trials using time manipulation is that the training was focusing on the failure states more than without time manipulation and managed to find ways to avoid them better. As a consequence, the speed of learning was increased and the algorithm learned earlier how to keep the balance longer, which lead to longer benchmark trials.

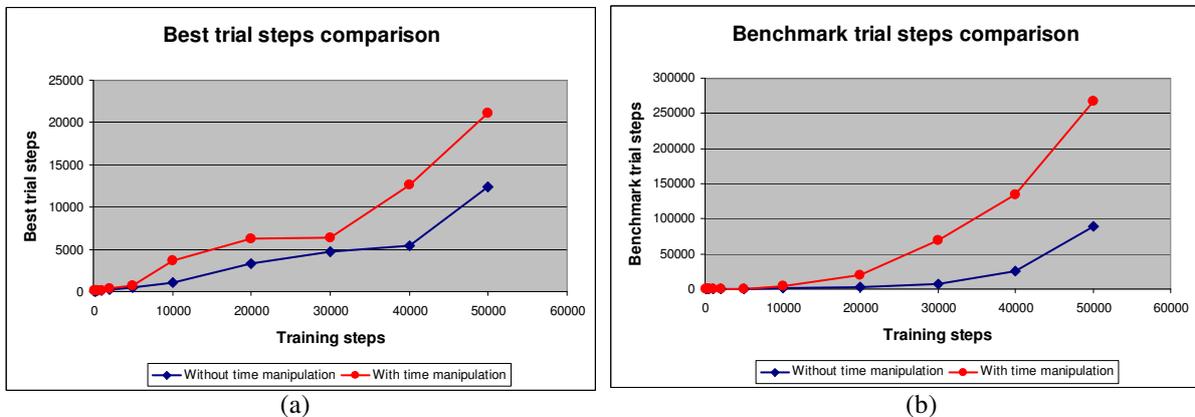

Figure 2. Comparing the learning performance with and without the time manipulation technique: (a) Best trial steps comparison and (b) Benchmark trial steps comparison.

An important observation is that the time manipulation technique has bigger impact on the speed of learning as the duration of the training increases. The reason for this is that the conventional algorithm "wastes" increasingly bigger amounts of steps going through the same state transitions from the beginning of the trial, which do not provide new experience for the learning algorithm to improve its policy. In addition, the time manipulation technique makes the learning algorithm focus on the problematic close-to-failure states and learn earlier a good policy to avoid them while at the same time avoiding repetition of already well-explored states. All these facts mean that the effect (i.e. increase of the learning speed) of the proposed technique is becoming bigger and bigger as the training duration increases. It starts around 20% for the shorter training sets and increases to more than 400% on the longer ones, resulting in a total averaged increase of about 260%. As a result, the speed of learning is increased by a factor of 2 on average.

4.3. Evaluation of the technique's effect on the state space exploration

In order to analyze the state space exploration, the entire continuous state space of the problem was divided in 162 discrete states. The division was based on dividing the cart track in 3 parts (left, center, right), dividing the angle of the pole in 6 parts (using 0, 1, 6 and 12 degrees for boundaries), dividing the linear velocity of the cart in 3 parts and the angular velocity also in 3 (3 x 6 x 3 x 3 = 162). Using this division, the state space exploration was easier to visualize and analyze.

For the purpose of visual inspection of the state space exploration, a special module in TiME was developed for displaying the state transitions graph in real-time during the learning trial. The module shows the state transitions graph using different graph layout algorithms, subgraph highlighting and graph operations animation for better visibility. During the experiments, this module proved to be an extremely valuable tool for evaluating the state space exploration.

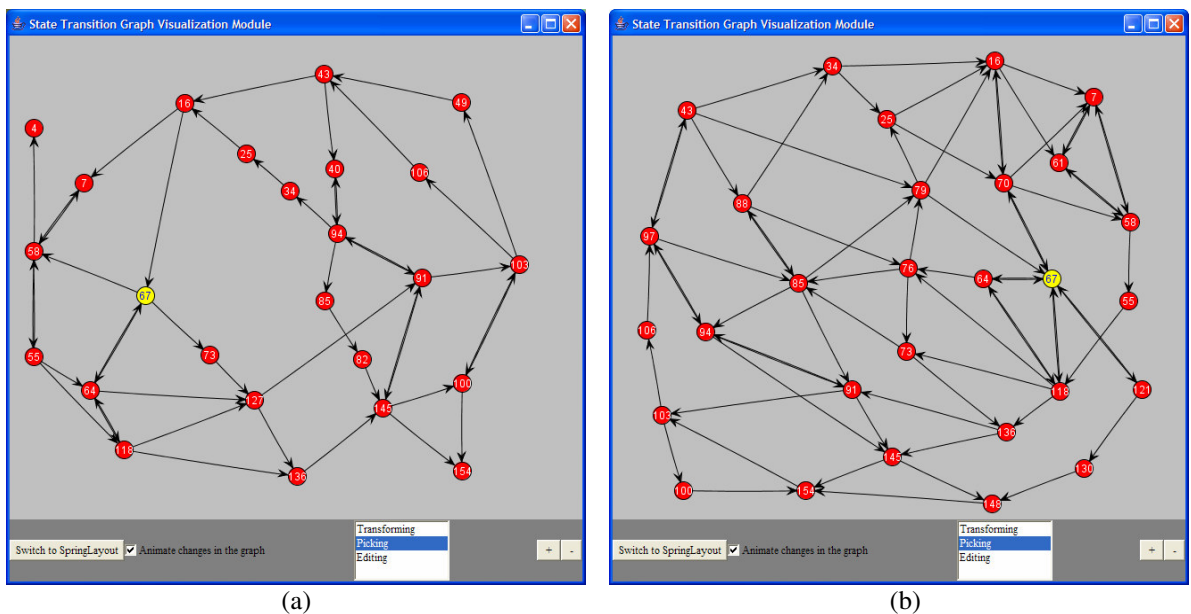

(a)  (b)
Figure 3. Comparing the state space exploration: (a) without time manipulation and
(b) with time manipulation.

Each state in the state transitions graph is represented by one numbered node, corresponding to the discretization of the continuous variables describing the state. The active state is marked with a yellow color, and the thickness of the edges shows the number of times the corresponding state transition occurred during the current learning trial. The developed graph visualization module allows transforming, picking and editing operations to be conducted on the graph for more detailed analysis.

As expected, the state space exploration of the proposed algorithm is broader than the conventional one. Figure 3 shows a visual comparison of the state transitions graph with and without the time manipulation technique after 300 steps of training. It is obvious that the proposed technique managed to explore more states (thus, more vertices in the graph) and to find bigger number of different state transitions (thus, better connected graph for the same number of edges).

In order to confirm objectively the visually observed results, 12 groups of experiments were conducted. The two algorithm versions (with and without time manipulation) were trained for the same number of increasing total steps. During the training, the number of unique (different) states visited was recorded. This number can justifiably be used as a measure of how good the state space exploration was during the training. The last two columns in Table 2 contain the experimental results and Figure 4 shows a comparison chart of the unique states visited data with and without the time manipulation technique.

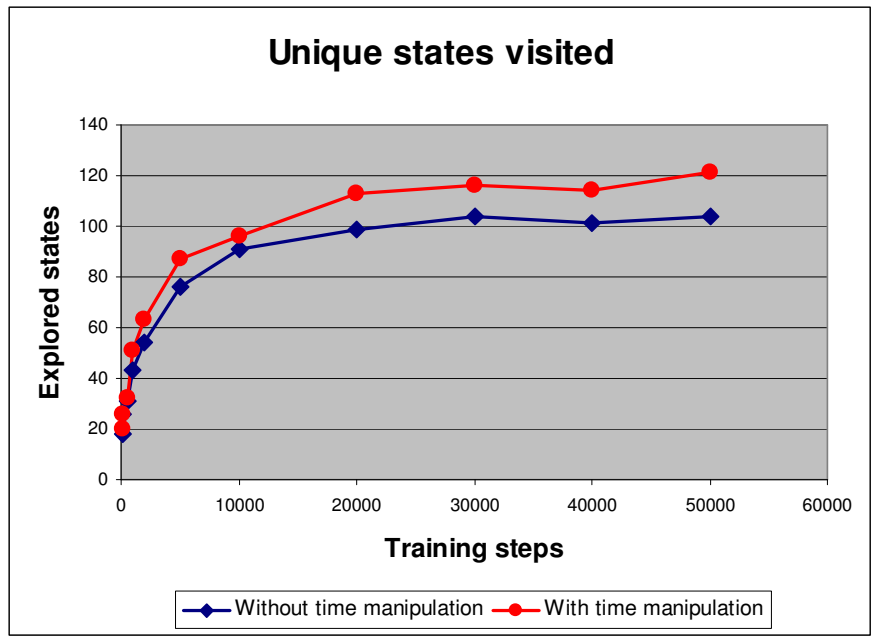

Figure 4. Comparing the state space exploration with and without the time manipulation technique.

Again, very promising results have been obtained. The state space exploration has improved as much as 19% in some of the longer training sets and 12% on average. The explanation for this result is that the proposed time manipulation technique makes the RL algorithm explore deeper the state space, thus discovering bigger number of different states and transitions between them. This has an additional positive effect on the speed of learning.

## 5. Conclusions and future perspectives

A time manipulation technique to improve the general RL algorithm has been proposed. It is applied to the standard Q-learning RL algorithm for solving failure-avoidance problems and thus a new Time Manipulated RL algorithm is created. The algorithm is tested on the classical pole-balancing task and it shows promising first results.

It is confirmed that the time manipulation technique improves the state space exploration by allowing the RL algorithm to explore better the state space in proximity to failure states. Also, it prevents the algorithm from "wasting" time to explore already well-explored areas of the state space. As a consequence, the RL agent is able to make much longer trials and explore the state space deeper compared with the unmodified RL algorithm for the same amount of simulation steps. It is shown that the proposed technique speeds up the learning by 260% and improves the state space exploration by 12% on average on the classical cart-pole balancing task, compared with the conventional Q-learning and Actor-Critic algorithms.

This research can be continued in many directions. For instance, the time manipulation technique can be tested with a wider range of failure-avoidance problems. It is interesting to see how it performs in the case of higher dimensional state and action spaces. Another possibility is to develop different strategies for choosing how many steps in the current trial to go back when a failure occurs. It could depend on local state information or global optimization parameters. It is also possible to devise other methods for selecting actions. For example, such methods could take into account the additional information from the time manipulations and try to make "smart selection" of actions based on this additional information.

In the future, a more generalized time manipulation framework can be developed. One possible generalization is to extend the scope of the time manipulations to include not only going backward, but also going forward in time. Such an audacious possibility is not completely absurd in a simulation environment, especially if multiple parallel simulations are executed simultaneously.

*Acknowledgements:* The authors would like to express their gratitude to Professor Gennady Agre from the Bulgarian Academy of Sciences for his valuable comments and suggestions.
This work is partially supported by MEXT (the Japanese Ministry of Education, Culture, Sports, Science and Technology).